# Single-view and Multi-view Fused Depth Estimation with Mamba

Zelin Meng and Zhichen Wang

## Abstract

Multi-view depth estimation has achieved impressive performance over various benchmarks. However, almost all current multi-view systems rely on given ideal camera poses, which are unavailable in many real-world scenarios, such as autonomous driving. In this work, we propose a new robustness benchmark to evaluate the depth estimation system under various noisy pose settings. Surprisingly, we find current multi-view depth estimation methods or single-view and multi-view fusion methods will fail when given noisy pose settings. To tackle this challenge, we propose a two-branch network architecture which fuses the depth estimation results of single-view and multi-view branch. In specific, we introduced mamba to serve as feature extraction backbone and propose an attention-based fusion methods which adaptively select the most robust estimation results between the two branches. Thus, the proposed method can perform well on some challenging scenes including dynamic objects, texture-less regions, etc. Ablation studies prove the effectiveness of the backbone and fusion method, while evaluation experiments on challenging benchmarks (KITTI and DDAD) show that the proposed method achieves a competitive performance compared to the state-of-the-art methods.

## Introduction

Depth estimation from images remains a critical challenge in computer vision with extensive applications, particularly in autonomous driving systems where understanding the 3D environment is crucial. Traditional methods for depth estimation are broadly categorized into multi-view and single-view approaches. Although recent single-view methods have achieved significant improvements in estimation accuracy, like diffusion model-based Marigold [5] and DepthAnything [4] which is trained by huge amount of training samples. Existing problems such as scale ambiguity still influence the upper-bound of estimation results. To overcome the disadvantages of single-view methods, some multi-view methods rely on epipolar geometry and triangulation, which necessitate precise camera calibration and accurate pose estimation. These methods, while providing high-quality depth information under ideal conditions, struggle with dynamic objects, texture-less regions, and scenarios where cameras are stationary, such as when vehicles stop at traffic lights or perform turns. Conversely, single-view methods offer robustness to these challenges by utilizing semantic understanding and perspective cues, yet they grapple with scale ambiguity, leading to inferior performance compared to multi-view methods.

Recent efforts to fuse these approaches have sought to leverage their respective strengths. However, existing fusion systems often assume ideal camera poses and can perform worse than single-view methods in the presence of noisy poses. To address these limitations, we propose a novel adaptive fusion network that integrates the benefits of multi-view and single-view depth estimation techniques while mitigating their disadvantages. Our approach utilizes a two-branch network, where each branch extracts image features using the designed Pyramid Mamba module. These extracted feature maps, which represent different scales, are then used to construct the corresponding cost volume. Following this, we perform feature fusion using the proposed Adaptive Fusion module. This module, featuring a cascade structure, fuses features at different scales in a coarse-to-fine manner to achieve accurate depth estimation. The main contributions of the proposed method can be summarized as follows:

- We introduce a single-view and multi-view fusion network where each branch utilizes the Pyramid Mamba module to extract multi-scale image features. These features are used to construct variance volume, which is then processed by the adaptive fusion module.
- We propose an adaptive fusion module which applies the calculated attention weights on variance volume, resulting in improved depth estimation accuracy even in challenging scenarios. Our approach overcomes the limitations of traditional methods by providing robust depth prediction results that is less affected by noisy camera poses and dynamic environments.
- We conduct extensive experiments and demonstrate various experimental results on benchmark datasets. And it shows our proposed method achieves a competitive performance on benchmark datasets including KITTI and DDAD.

**Related Works**

**Single-view Based Methods**

Single-view depth estimation remains a central challenge in computer vision, with recent advancements largely driven by convolutional neural network (CNN)-based approaches. Traditionally, these methods tackle the problem through per-pixel classification or regression frameworks. Eigen et al. [6] firstly proposed a two-stage neural network-based model, which shows the potential of CNN-based methods on tasks of single-view depth estimation. To enhance performance, recent strategies have included aggregating more robust visual features, developed novel loss functions, and utilized mix-data training techniques. [12] presented a compact but effective model for self-supervised single-view depth estimation, which somehow overcomes the limitation of obtaining accurate per-pixel ground-truth depth data. In [17], the authors proposed a feature metric loss which enables the learning of feature representations in a self-supervised manner. To reduce the increasing complexity of depth estimation networks, [8] reformulate the training of depth estimation network as an ordinal regression problem, and adopt a muti-scale network structure to avoid the unnecessary spatial pooling. [12] propose a network architecture that utilizes novel local planar guidance layers to enhance the performance of encoder-decoder structure on depth prediction tasks. [2] proposed a transformer-based block to that divides the depth range into bins whose center value is estimated adaptively per image, which leads to a significant improvement in monocular depth prediction. [22] propose a canonical camera space transformation module which enables monocular models to be trained in an extremely mixed-large dataset and obtaining excellent zero-shot generalization abilities. Despite these innovations and continuous improvements on various benchmarks, the accuracy achieved by single-view depth estimation methods still lag behind that of multi-view geometry-based methods. Considering this, our work integrates a single-view depth estimation module, leveraging its notable robustness to low-texture regions and dynamic objects, to address some of the limitations inherent in current approaches.

**Multi-view Based Methods**

Multi-view depth estimation has significantly advanced through leveraging multiple camera views with known intrinsics and poses. Initial approaches, such as those introduced by [24], demonstrated the potential of feature learning in multi-view stereo, albeit using traditional aggregation methods for matching costs. Building on this, [MVSNet] pioneered the use of a differentiable cost volume combined with 3D CNNs for cost regularization, achieving the leading accuracy of its time. Contemporary state-of-the-art methods largely adhere to this paradigm [3, 23, 26, 51]. [19] propose a novel and learnable cascade formulation of Patch-match for high-resolution multi-view stereo. [21] present a highly efficient multi-view stereo algorithm that seamlessly integrates multi-view constraints into single-view networks via an attention mechanism, which significantly reduce the computation burden. [20] propose a GRU-based estimator to encode pixel-wise probability distribution of depth. [11] propose a memory and time efficient cost volume formulation. [1] fuses single-view depth probability with multi-view geometry, to improve the accuracy, robustness and efficiency of multi view depth estimation. Some other methods tackles the depth estimation task from another perspective, [18] combines the representation ability of neural networks with the geometric principles governing image formation. However, these approaches face limitations, including a heavy reliance on high-parallax motion and diverse textures for optimal performance, challenges in handling dynamic objects, and sensitivity to inaccuracies in pose estimation, particularly when dealing with pose noise.

The fusion of single-view and multi-view depth predictions offers a promising approach to leverage the strengths of both methods, aiming to enhance accuracy and efficiency. Several methods have been proposed in this domain. For instance, [7] improves accuracy by integrating the local details from single-view depth estimates with the robust multi-view predictions in regions of high parallax and gradient. [1] reduces computational overhead by first predicting depth through a single-view network and then constructing a thin cost volume, achieving better accuracy with lower computational demands. However, these approaches face a critical challenge: their reliance on accurate pose information. Inaccuracies in pose estimation can severely degrade performance, sometimes making these methods less effective than single-view approaches. [21] addresses this issue with an Epipolar Attention Module that fuses single-view and multi-view data while incorporating attention mechanisms at various resolutions.

Although this method offers some improvement in handling pose noise, it still struggles with integrating incorrect matching information from noisy poses, which can undermine overall accuracy. Thus, we propose a two-branch network architecture that incorporated with adaptive fusion module to combine the reliable results from the two branches, which achieves a competitive estimation performance while significantly enhances the robustness of system.

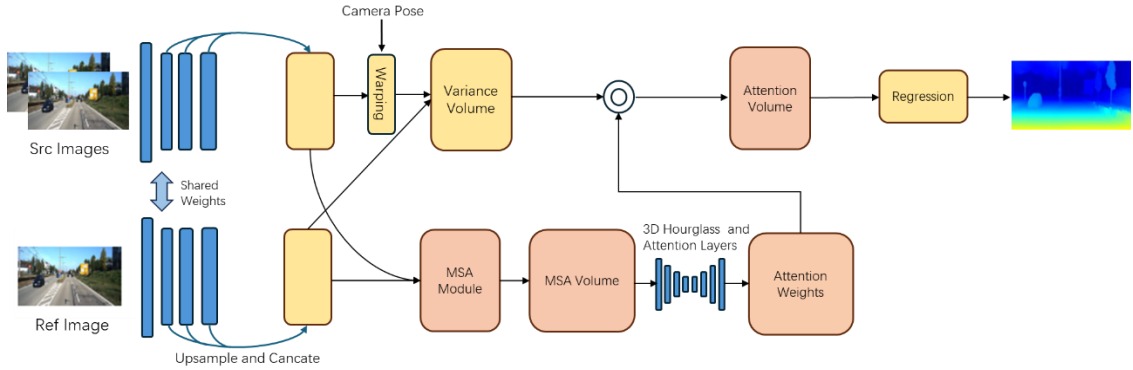

Figure. 1 Overall pipeline of the proposed method, which consists of three parts: multi-scale feature extraction, attention volume construction and depth regression. Single-view branch and multi-view branch share the feature extraction backbone, extracted features are utilized to construct variance volume and attention volume. The final depth map is obtained through a residual regression network.

## 3. Proposed Methods

The primary works of the proposed method is to perform depth estimation from images using a multi-view stereo approach within a neural network framework. The forward pass extracts features from both a reference image and multiple source images, constructs a 3D volume representation based on these features, and calculates depth predictions using attention mechanisms and regression techniques. The code facilitates the modeling of depth across multiple perspectives, leveraging convolutional neural networks to manage both feature extraction and the depth regression process, ultimately outputting depth maps along with their confidence scores. The network is designed with a modular architecture that incorporates multiple layers to extract features at various scales. Each layer is responsible for capturing different levels of detail, from low-level textures to high-level semantic information. Utilizing Vmamba, the network integrates multiple scales of feature extraction. This is achieved through parallel paths that process the input at different resolutions, allowing for a more comprehensive understanding of the data. The architecture can include convolutional layers, pooling layers, and possibly attention mechanisms that enhance the focus on relevant features. The design is flexible, allowing for adjustments in layer depth and width to suit different datasets or tasks.

## 3.1 Feature Extraction

The proposed method for depth estimation utilizes a multi-view stereo approach which is implemented by a two-branch neural networks. The reference image is processed through a feature extractor to obtain a series of multi-scale feature maps.

The network is designed with a modular architecture that incorporates multiple layers to extract features at various scales. Each layer is responsible for capturing different levels of detail, from low-level textures to high-level semantic information. Utilizing proposed depth-mamba backbone, the network integrates multiple scales of feature extraction. This is achieved through parallel paths that process the input at different resolutions, allowing for a more comprehensive understanding of the data [25, 26]. The architecture includes a stack of depth mamba modules that can enhance the focus on relevant features. The design is flexible, allowing for adjustments in layer depth and width to suit different datasets or tasks. Besides, inspired by [26], we also propose local feature blocks to further enhance the ability of extracting local features because mamba is not good at capturing the spatial information of input images compared to other CNN-based methods.

## 3.2 Attention Volume Construction

In our proposal, we construct attention volume for the depth regression. For the branch of reference image, we firstly construct the group-wise correlation (GwC)volume by utilizing both reference features and source features captured through depth-mamba backbone. After that, we apply a multi-scale attention module and a 3D hourglass network on the GwC volume to obtain attention weights. For the branch of source images, we construct variance volume by utilizing the reference feature and source feature. Finally, we multiply the constructed variance volume and attention weights to obtain the attention volume.

## 3.3 Depth Value Regression

Depth predictions are generated through a depth value regression process applied to the feature maps, resulting depth maps and confidence maps. Finally, the depth and confidence maps are up-sampled to the original input resolution using nearest neighbor interpolation, and the results are organized into the outputs dictionary, which encapsulates the method's ability to deliver accurate depth estimation from multi-view images.

## 3.4 Loss Functions

In the training process, we adopt mean absolute error (MAE) loss function as it is commonly used in regression tasks to measure the accuracy of a model. The loss function is defined as follows:

$$L = \frac{1}{N}\sum_{i=1}^{N}|y_i - \hat{y}_i| \qquad (1)$$

Where N denotes the total number of pixels, $y_i$ indicates the ground truth depth value for the $i_{th}$ pixel, $\hat{y}_i$ indicates the predicted depth value for $i_{th}$ pixel.

## 4. Experiments

### 4.1 Dataset

In stereo depth estimation task, high-quality datasets are significantly essential for training and evaluating depth estimation models. Two prominent datasets in this field are the KITTI and DDAD datasets, each providing valuable resources for developing and benchmarking stereo vision systems.

The KITTI dataset, introduced by the Karlsruhe Institute of Technology and Toyota Technological Institute, is one of the most widely used benchmarks for stereo depth estimation. It offers a diverse set of stereo image pairs captured from a vehicle-mounted stereo camera rig in various urban and rural environments. The KITTI dataset includes a range of scenes, from city streets to highways, with annotations for ground truth disparity maps and 3D point clouds. This dataset is renowned for its challenging conditions, including varying lighting, weather, and dynamic objects, making it a robust testbed for evaluating depth estimation algorithms. The image resolution of stereo pairs on KITTI is around 1241×376. And we train and test the proposed method on KITTI Eigen split [6].

The DDAD (Driving Dataset for Autonomous Driving) dataset is a more recent addition designed to address specific challenges in stereo depth estimation for autonomous driving. Collected using a stereo camera system mounted on a vehicle, the DDAD dataset includes high-resolution stereo image pairs and depth annotations, with a focus on dynamic and complex driving scenarios. The DDAD dataset emphasizes conditions such as dense traffic, diverse weather, and varying road textures, providing a rich source of data for developing and testing depth estimation methods under real-world driving conditions. In specific, DDAD dataset utilized in this study is captured using six synchronized cameras, providing a comprehensive 360-degree field of view. It features high-accuracy ground-truth depth information obtained from high-density LiDARs, ensuring precise depth measurements. The dataset comprises 12,650 training samples and 3,950 validation samples, each derived from a single camera view with a resolution of 1936×1216 pixels. Both the individual camera views and the integrated data from all six cameras are employed for training and testing, offering a robust foundation for evaluating depth estimation methods across diverse scenarios.

## 4.2 Evaluation Metrics

The evaluation metrics used for quantitative evaluation are defined as follows:

$$AbsRel = \frac{1}{N}\sum_{i=1}^{N} \left|\frac{y_i - y_i^{pred}}{y_i}\right| \qquad (2)$$

$$SqRel = \frac{1}{N}\sum_{i=1}^{N} \left(\frac{y_i - y_i^{pred}}{y_i}\right)^2 \qquad (3)$$

$$RMSE = \sqrt{\frac{1}{N}\sum_{i=1}^{N} \left(y_i - y_i^{pred}\right)^2} \qquad (4)$$

Where $N$ denotes the total number of pixels with ground truth labels in the depth map.

$y_i$ indicates the ground truth depth value and $y_i^{pred}$ represents predicted depth value for $i_{th}$ pixel. The RMSE measures the average magnitude of the error between predicted and actual depth values. The AbsRel metric measures the mean of the absolute relative errors. The SqRel metric measures the mean of the squared relative errors.

Besides, we also adopt delta metrics to verify the performance of our proposed methods. In specific, for each pixel $p$, compute the ratio as follows:

$$r(p) = \min\left(\frac{\widehat{D}(p)}{D(p)}, \frac{D(p)}{\widehat{D}(p)}\right) \quad (5)$$

Then, we count the number of pixels where $r(p) < 1.25$ for $\delta_1$, $r(p) < 1.25^2$ for $\delta_2$, $r(p) < 1.25^3$ for $\delta_3$, respectively. Finally, we divide these counts by the total number of pixels to get the percentage of pixels meeting the criteria for each metric.

### 4.3 Implementation Details

We implement the proposed methods with PyTorch [16] and conduct evaluation experiments utilizing NVIDIA RTX 4090 GPUs. We adopt AdamW optimizer [14] and schedule the learning rate utilizing one-cycle policy [31] with lrmax = 1.0× 10−4. We trained 40 epochs on KITTI [6] and 30 epochs on DDAD [10]. During the training process, the proposed network architecture takes consecutive 3 frames as input, i.e, n=3. Regarding the details of multi-view setup, we set the depth hypothesis number of planes as 128, besides, the weight of the frames from different views is the same.

### 4.4 Comparison Experiment Results

To reveal the competitive performance of the proposed methods, we conduct evaluation experiments on KITTI [6] and DDAD [10] benchmark.

The KITTI dataset [9] and the KITTI Eigen split [6] serve as key benchmarks for evaluating both single-view and multi-view depth estimation. We present a comparison with state-of-the-art techniques on this benchmark, with results shown in Table 1. Our proposed model achieves an AbsRel error of 0.042, a highly competitive performance compared to the recent approaches. Notably, our proposal reduces the RMSE error by 2.8% compared to the current state-of-the-art method AFNet [2].

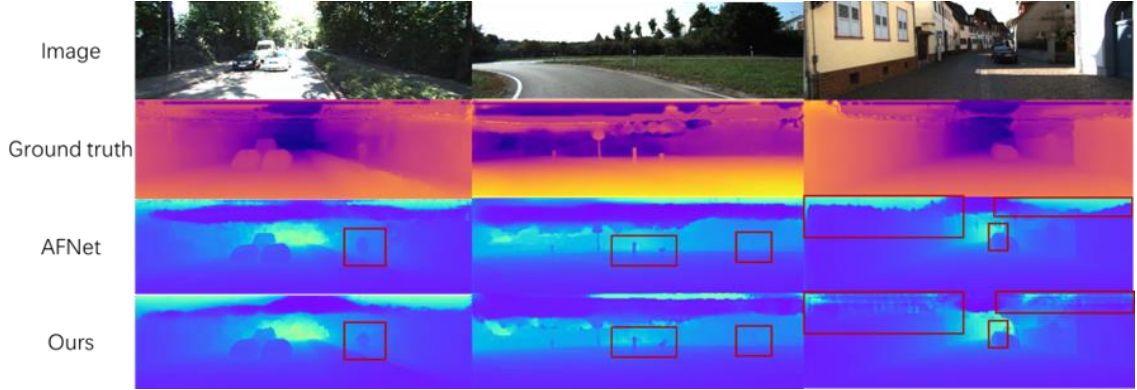

Figure. 2 Qualitative results on KITTI benchmark. Red rectangles highlight obvious enhancements achieved by our proposed method.

The DDAD dataset [10] has not been widely used for training and testing by most traditional multi-view and single-view methods. Therefore, we apply the same training procedure across all methods (refer to Section 4.2 for more details). It is important to note that all methods have successfully converged. The quantitative results are presented in Table 1, where our proposed method achieves competitive performance compared to AFNet on DDAD. Compared to the current SOTA methods AFNet and MgaNet [1], our proposal shows an improvement of 15.3% in RMSE error. Qualitative results are displayed in Figure 3, where our proposal outperforms in static objects and the background area.

## 4.5 Ablation Studies

To demonstrate the effectiveness of each component of the proposed method, we conducted ablation studies on KITTI. In this section, the following variants are discussed to evaluate the effectiveness of the proposed method:

### 4.5.1 Ablation Studies on Feature Extraction Backbone

Based on the two-branch multi-view depth estimation architecture, we employ three different kinds of backbone, including ConvNeXt-T, VMamba, and our proposed Depth-mamba to verify the effectiveness of the proposed backbone towards this depth estimation task.

The results of ablation experiments are summarized in Table.2. The quantitative results reveal the effectiveness of our proposed mamba-based backbone. The proposed backbone outperform the ConvNeXt-T and VMamba.

### 4.5.2 Ablation Studies on Fusion Module

To evaluate the effectiveness and performance of the proposed fusion module, we conduct ablation studies on KITTI. Based on the basic architecture of two-branch muti-view depth estimation architecture, we implemented three variants. For the case of directly concatenating, we directly concatenate the two cost volumes utilizing a convolution layer which are constructed on single-view branch and multi-view branch, respectively. For the case of utilizing cross attention, we apply a cross-attention layer to fuse the cost volumes that are constructed on single-view and multi-view branch. For the details of our proposed fusion method, please refer to section 3.2.

**Ablation on proposed backbone.** The comparison results are shown in Table 2. Compared with the commonly adopted 'ConvNeXt-T', applying the proposed depth-mamba can extract more robust features from the two-branch inputs leading to the AbsRel error of the depth prediction reduced by 27.5%. Compared to VMamba backbone, the proposed method reduces the AbsRel error by 22.2%, which reveals the competitive performance of the proposed fusion method.

Effectiveness of the proposed feature extraction backbone. As shown in Table 2, the evaluation metrics indicate the effectiveness and robustness of the proposed depth-mamba for tackling the task of multi-view depth prediction.

**Ablation on proposed fusion.** The comparison results are shown in Table 3. Compared with 'Concatenation', applying the attention weights on the variance volume and supervising the output can extract more robust features, leading to the AbsRel error of the depth prediction ('Concatenation' v.s. 'Ours') reduced by 33.3%. Compared to 'Cross-attention', the proposed fusion method reduces the AbsRel error by 23.63%, which reveals the competitive performance of the proposed fusion method.

Effectiveness of the Adaptive Fusion. As shown in Table 3, the accuracy of the proposed fusion method 'Ours' is significantly higher than 'Concatenation' since the latter method of direct convolution fusion was quite naive. Thus, we propose a fusion module to replace this crude way of fusion. Comparing 'Ours' with 'Concatenation', using the proposed fusion module achieves a significant reduction on AbsRel error.

Table 2.

| Model | AbsRel↓ | SqRel↓ | RMSE↓ |
|---|---|---|---|
| Base-Mamba | 0.058 | 0.195 | 2.216 |
| Base-VMamba | 0.054 | 0.189 | 2.175 |
| Base-DepthMamba | 0.042 | 0.121 | 1.695 |

Table 3.

| Fusion Module | AbsRel↓ | SqRel↓ | RMSE↓ |
|---|---|---|---|
| Concatenation | 0.062 | 0.201 | 2.517 |
| Cross-Attention | 0.055 | 0.193 | 2.198 |
| Proposed-Fusion | 0.042 | 0.121 | 1.695 |

Table 1.

| Type | Method | DDAD | KITTI |
|---|---|---|---|

|  |  | AbsRel↓ | SqRel↓ | RMSE↓ | AbsRel↓ | SqRel↓ | RMSE↓ |
|---|---|---|---|---|---|---|---|
| Single-View | Monodepth2 | 0.194* | 3.52* | 13.32* | 0.106 | 0.806 | 4.630 |
|  | FeatDepth | 0.189* | 3.21* | 12.45* | 0.099 | 0.697 | 4.427 |
|  | DORN | - | - | - | 0.088 | 0.806 | 3.128 |
|  | BTS | 0.169* | 2.81 | 11.85* | 0.059 | 0.245 | 2.756 |
|  | AdaBins | 0.164* | 2.66* | 11.08* | 0.058 | 0.190 | 2.360 |
|  | Metric3D | 0.183* | 2.92* | 12.15* | 0.053 | 0.174 | 2.243 |
| Multi-View | PMNet | 0.141* | 2.23* | 10.56* | - | - | - |
|  | Deepv2d | - | - | - | 0.091 | 0.582 | 3.644 |
|  | CasMVS | 0.129* | 2.01* | 9.87* | 0.066* | 0.228* | 2.567* |
|  | MVSNet | 0.109* | 1.62* | 8.21* | - | - | - |
|  | IterMVS | 0.104* | 1.59* | 7.95* | 0.057* | 0.178* | 2.234* |
|  | MVS2D | 0.132* | 2.05* | 9.82* | 0.058* | 0.176* | 2.277* |
|  | SC-GAN | - | - | - | 0.063 | 0.178 | 2.129 |
|  | MaGNet | 0.112* | 1.74* | 9.23* | 0.054 | 0.162 | 2.158 |
|  | AFNet | 0.088 | 1.41 | 7.23 | 0.044 | 0.121 | 1.743 |
|  | Ours | 0.107 | 1.49 | 6.12 | 0.042 | 0.121 | 1.695 |

## Conclusions

In this paper, we propose a two-branch neural networks which fuses the most reliable depth estimation results obtained from the single-view and multi-view branch, to improve the robustness of depth estimation results under the challenging conditions. In specific, we select the most robust depth estimation results from the two branches by utilizing an attention-based method. Besides, we introduce the mamba-based feature extraction backbone to further enhance the feature representation ability of the overall networks. The ablation studies certificate the effectiveness of the proposed fusion method and mamba-based backbone. Evaluation experiments show the competitive performance of our proposed method on both DDAD and KITTI benchmarks.

arXiv preprint arXiv:2312.00752, 2023. 12